# Automated Diagnosis of Epilepsy Employing Multifractal Detrended Fluctuation Analysis Based Features

S. Pratiher[1], S.Chatterjee[2] and R.Bose[3]
[1] Department of Electrical Engineering, Indian Institute of Technology, Kanpur, India
[2] Electrical Engineering Department, Jadavpur University, Kolkata, India
[3] Electrical Engineering Department, Calcutta Institute of Engineering and Management, Kolkata, India
{ sawon1234, chapeshwar, rohitbose94 } @gmail.com

**Abstract:**
This contribution reports an application of Multi Fractal Detrended Fluctuation Analysis (MFDFA) based novel feature extraction technique for automated detection of epilepsy. In fractal geometry, Multi-fractal Detrended Fluctuation Analysis (MFDFA) is a popular technique to examine the self-similarity of a non linear, chaotic and noisy time series. In the present research work, EEG signals representing healthy, interictal (seizure free) and ictal activities (seizure) are acquired from an existing available database. The acquired EEG signals of different states are at first analyzed using MFDFA. To requisite the time series singularity quantification at local and global scales, a novel set of fourteen different features. Suitable feature ranking employing student's *t*-test has been done to select the most statistically significant features which are henceforth being used as inputs to a support vector machines (SVM) classifier for the classification of different EEG signals. Eight different classification problems have been presented in this paper and it has been observed that the overall classification accuracy using MFDFA based features are reasonably satisfactory for all classification problems. The performance of the proposed method are also found to be quite commensurable and in some cases even better when compared with the results published in existing literatures studied on the similar data set.

## 1. Introduction

Epilepsy is a very serious neurological disorder which affects about 1%-2% of human population on earth. Epilepsy is generally characterized by recurrent seizures which occur due to malfunctioning of neurons located inside the human brain [1]. During seizure activity, the electrical signals transmitted by neurons become highly abnormal in nature. Symptoms like severe jerking movements of legs, arms, loss of consciousness and awareness etc. are very common among the patients suffering from epilepsy. Early and accurate detection of epilepsy is, therefore exigent to prevent these unusual and undesirable physiological abnormalities. In pathology labs and clinics, epilepsy is detected usually by expert neurologists through electroencephalogram (EEG) screening, owing to the fact that EEG based diagnosis is inexpensive and has better time resolution compared to fMRI-based treatment [2]. But the detection of epilepsy through visual inspection of EEG recordings, are often incorrect and inevitably lengthy. Hence, an automated computer aided fast and accurate disease detection scheme to detect epilepsy seizures correctly and at a much lesser time have become an utmost necessity. Considering the above-said fact, automatic detection of epilepsy using suitable signal processing and machine learning algorithms have therefore been a major focal point of research over the last couple of years.

Analysis of epileptic seizure and healthy EEG signals in time domain using cross-correlation and Least Square Support Vector machines was reported by Chandaka et al. in [3]. In Frequency domain, spectral analysis based on Fast Fourier Transform and Decision Tree classifier was employed for automatic detection and classification of epileptic seizures [4]. Artificial neural network based combined time and frequency domain features have also been successfully implemented to detect epileptic seizures in EEG signals in [5]. Analysis of EEG signals in joint time frequency domain, based on wavelet transform and mixture of expert model have been reported in many existing literatures [6]. Epileptic seizure detection based on multiwavelet transform and approximate entropy employing Artificial Neural Networks have been reported in [7]. Analysis of seizure and seizure free EEG signals based on empirical mode decomposition have been reported in many available literatures. Several feature parameters including, instantaneous area, second order difference plot, bandwidth features, phase space reconstruction etc. derived from respective intrinsic mode functions (IMF's) have been used as inputs to a Least square Support Vector Machines (LS-SVM) classifier for classification of EEG signals [8-11]. Feature parameters based on local binary patterns, key point local binary patterns using LS-SVM classifier for automated detection of epilepsy have been reported in [12-13]. Since, EEG signals are representatives of complex brain dynamics, they manifest highly non linear and chaotic behaviour. Therefore, analysis of EEG signals based on several non linear techniques like Approximate Entropy, Fractal Dimension, Lyapnov exponent etc. for the purpose of detection and classification of epileptic seizures in EEG signals have been reported in [14-16]. Detection of epilepsy based on weighted visibility graph based features and fractal dimension of Flexible Analytic Wavelet Transform have been very reported very recently in [17-18]. Hence, it is evident from the existing literature survey, that EEG signals are typically manifest non stationary and non linear behavior. Therefore, non linear signal processing techniques can be effectively applied for analysis and classification of different categories of EEG signals. Considering the above fact, in this contribution, feature parameters based on non linear analysis of EEG signals employing Multi Fractal Detrended Fluctuation Analysis (MFDFA) have been reported for discrimination of different types of EEG signals.

Detrended Fluctuation Analysis (DFA) was first proposed by Peng et.al. to detect the long range correlation of DNA sequences [19]. Since then, DFA has been used widely for the determination of monofractal scaling properties in noisy, non stationary time series [20-21]. However, many real life signals do not exhibit simple monofractal behaviour i.e. they cannot be characterized by a single scaling exponent, rather different scaling exponents are required to manifest the characteristic of different parts of a non linear time series. This led to the development of Multi Fractal Detrended Fluctuation Analysis (MFDFA). MFDFA was first conceived by Kantelhardt et al. [22] as a generalization of the standard Detrended Fluctuation Analysis (DFA). MFDFA overcame the limitation of a single scaling exponent of conventional DFA method using different order fluctuation functions. Using fluctuation functions of different order, the scaling behaviour of a nonlinear time series can be analysed in different segments. Another distinct advantage of using MFDFA technique is that it has low computational burden compared to existing techniques like Wavelet transform maximum modulus (WTMM) for determining the long range correlation of a multifractal time series [22]. MFDFA has been applied successfully to study the non linear and chaotic nature of various time series like partial discharge signals [23], bearing fault signals [24] and also for analysis of several physiological signals including EEG [25]. This work not only deals with the analysis of EEG signals based on MFDFA, but also classification of EEG signals based on several new features extracted from multi-fractal spectrum (MFS) of EEG signals have also been presented, which has not been reported so far in any literatures. In this contribution, fourteen different feature parameters obtained from the respective MFS of different EEG signals are being used for effective discrimination of different EEG signals. After feature ranking using student's t-test, four highly discriminative features are selected which are henceforth used as inputs to SVM and kNN classifiers for the purpose of classification of EEG signals.

The paper is divided into following sections. Section II explains the EEG data set used in this work following by the brief steps of MFDFA in section III. Section IV deals with the extracted features and their physical significance. Section V provides a brief theory of SVM classifier. Finally, results and discussion are given in section VI, followed by conclusion in section VII.

## 2. EEG Signal Data set

In the present work, EEG signals are taken from online available benchmark database of University of Bonn, Germany [26]. The dataset comprises of five sets of single channel EEG recordings denoted by A, B, C, D, and E. Length of each signal is 23.6 sec. The data is sampled at a sampling frequency of 173.61 Hz. Recording of EEG signals are done using similar 128-channel amplifier system, using an average common reference After the signals are recorded, band-pass filtering was performed with filter settings between 0.53–40 Hz. Each data set contains 100 single channel EEG segments. EEG signals of sets A and B are acquired from surface electrode placement using the standard 10-20 electrode system from five healthy volunteers in eyes open and eyes closed conditions respectively. EEG signals of sets C are recorded from the hippocampal formation in the opposite hemisphere and that of set D are recorded from the epileptogenic zone, respectively. Both sets C and D comprises of activity in the seizure free intervals whereas, Set E consists of seizure activities only. In the present study, eight different classification problems are presented by combining five different sets (A,B, C, D and E) of EEG signals. The different CP along with brief descriptions are presented in Table-I.

Table-1: Types of classification problem

| Classification Problem (CP) | Class | Description |
|---|---|---|
| I | A, E | Healthy with eyes open vs Seizure |
| II | B, E | Healthy with eyes closed vs Seizure |
| III | C, E | Hippocampal Interictal vs Seizure |
| IV | D, E | Epileptogenic Interictal vs Seizure |
| V | AB,E | Healthy vs Seizure |
| VI | CD,E | Interictal vs seizure |
| VII | AB,CD | Healthy vs Interictal |
| VIII | ABCD,E | Seizure free vs seizure |

## 3. Multi-Fractal Detrended Fluctuation Analysis

After acquisition of EEG signals of different sets, they are at first analysed using MFDFA. The basic steps of MFDFA as are described briefly as follows:

Let us consider a non stationary and non-linear time series $x(n)$ for n=1,……. N of length N.

*Step 1:* First step is to compute the mean of the time series given by

$$\bar{y} = \frac{1}{N}\sum_{n=1}^{N} y(n) \quad (1)$$

*Step 2:* After computation of mean value, in the next step, $\bar{y}$ is subtracted from the signal to compute the integrated time series given by

$$Y(i) = \sum_{n=1}^{i}[y(n) - \bar{y}] \quad \text{for} \quad i=1,\ldots,N \quad (2)$$

*Step 3:* The integrated time series is divided into $N_s$ number of non overlapping segments (where $N_s = \text{int}(N/s)$ and s is the time scale or length of each segment). When N is not a multiple of s, some data remains at the end of the series $Y(i)$. In order to include the remaining part of the series, the entire process is repeated again from the opposite end, thus giving a total number of $2N_s$ segments. The local RMS variation/trend of each segment out of total $2N_s$ segments is obtained by using a least square polynomial fit of the time series and then the variance for each segment is determined by

$$\Im^2(s,v) = \frac{1}{s}\sum_{i=1}^{s}\{Y[(v-1)s+i] - y_v(i)\}^2 \quad \text{for } v=1,\ldots,N_s \quad (3)$$

$$\Im^2(s,v) = \frac{1}{s}\sum_{i=1}^{s}\{Y[N-(v-N_s)s+i] - y_v(i)\}^2 \quad \text{for } v = N_s+1,\ldots,2N_s \quad (4)$$

Here $y_v(i)$ is the least square fitted value in the segment v.

*Step 4*: The $q^{th}$ order fluctuation function $\Im_q(s)$ is obtained after computing average procedure of $2N_s$ segments, where q is an index which can take all possible values except q=0. For q=0, a logarithmic averaging procedure is followed, instead of normal averaging procedure.

$$\Im_q(s) = \left\{\frac{1}{2N_s}\sum_{v=1}^{2N_s}[\Im^2(s,v)]^{0.5q}\right\}^{1/q} \quad (5)$$

From Equation (5), it is evident that the fluctuation function $F_q(s)$ depends on the time scale s for different values of q. The steps 2-4 are therefore repeated by varying the time scales s. For q=2, the method reduces to standard DFA.

*Step 5*: Variation of $\Im_q(s)$ versus s for each value of q is analysed using a log-log plot. When the analysed time series y(n) is long-range power-law correlated, variation of $\Im_q(s)$ versus s, shows a power-law behaviour, with h(q) as the slope, where h(q) is known as the generalized Hurst exponent, which depends on the value of q.

$$\Im_q(s) \alpha\ s^{h(q)} \quad (6)$$

For a monofractal time series, scaling behaviour of $\Im^2(s,v)$ is almost identical in all segments v, for all values of q. Hence, h(q) is independent of q. However, for a multifractal time series, h(q) is a function of q, and for q = 2, i.e. h(2) gives the value of simple Hurst exponent. The average value of $\Im_q(s)$ in equation (5) will be mainly influenced by larger and smaller variance of $\Im^2(s,v)$ within segment v, corresponding to q > 0 and q < 0, respectively. Therefore, h(q) describes the scaling behaviour of the segments with large and small fluctuations, respectively for positive and negative values of q. Further, the large fluctuations are generally characterized by a smaller scaling exponent h(q) for multifractal series and vice-versa [24-26].

*3.1 Determination of Multi-Fractal Spectrum*

The relationship between the generalized Hurst exponent h(q) of MFDFA and multifractal scaling exponent τ(q) is given by

$$\tau(q) = qh(q) - 1 \quad (7)$$

A monofractal time series with long range correlation is characterized by a single Hurst Exponent, where the multifractal scaling exponent τ(q) shows a linear dependency on q. On the contrary, multifractal time series have multiple Hurst exponents and τ(q) depends nonlinearly on q [50]. Using a Legendre transform, the relationship between the singularity spectrum f (α) and scaling exponent τ(q) is obtained, which are given by [23-25]

$$\alpha = \frac{d\tau}{dq} \quad (8)$$

$$\text{and } f(\alpha) = q\alpha - \tau(q) \quad (9)$$

where α is known as singularity exponent and f(α) is the fractal dimension of subset of the series characterized by α. Now, Using Eq. (6) α and f (α) can be expressed in terms of h(q) as follows

$$\alpha = h(q) + qh'(q) \quad (10)$$

$$f(\alpha) = q[\alpha - h(q)] + 1 \quad (11)$$

In general, the singularity spectrum f(α) quantifies the long range correlation property of a time series [24-26]. The shape of the multifractal spectrum looks like an inverted parabola, where the width of the parabola is a measure of the multifractality of the spectrum. A larger spectral width is an indicator of high degree of multifractality. For a monofractal time series, since h(q) is independent of q, the width will be zero.

## 4. Feature extraction using MFDFA

*4.1 Extracted Features:*

In the present work, EEG signals of five different sets representing different states of human brain are at first characterized by multifractal parameters. From the multifractal spectrums i.e. f (α) and α curves for five EEG signals, fourteen distinct new features are obtained for discrimination of different EEG signals. The proposed set of features which are used in this work are as follows:

$F_1$: Generalized Hurst Exponent

$F_2$: Singularity exponent (α) corresponding to peak of singularity spectrum=$α_{peak}$

$F_3$: Right extremity of the singularity exponent= $α_{max}$

$F_4$: Left extremity of the singularity exponent=$α_{min}$

$F_5$: Mean singularity exponent=$α_{mean}=(α_{max} + α_{min})/2$

$F_6$: Singularity spectrum width=$Δα=(α_{max} -α_{min})$

$F_7$: Horizontal distance between the peak and the minimum value of singularity exponent = $(α_{peak}-α_{min})$

$F_8$: Horizontal distance between the peak and the maximum value of singularity exponent = $(α_{peak}-α_{max})$

$F_9$: Singularity spectrum corresponding to $α_{max}$ =$f(α_{max})$

$F_{10}$: Singularity spectrum corresponding to $α_{min}$=$f(α_{min})$

$F_{11}$: Mean singularity spectrum=$f(α_0)$= $\{f(α_{max})+f(α_{min})\}/2$

$F_{12}$: Vertical distance between $f(α_{max})$ and $f(α_{min})$ =$Δf(α)$= $f(α_{max})- f(α_{min})$

$F_{13}$: Vertical distance between the peak value of singularity spectrum and $f(α_{min})$=$f(α_{peak})- f(α_{min})$

$F_{14}$: Vertical distance between the peak value of singularity spectrum and $f(α_{max})$= $f(α_{peak})- f(α_{min})$

*4.2 Physical significance of the extracted features:*

The significance of these features are explained briefly. $F_1$ is the generalized hurst exponent i.e. h(2) which indicates the long range autocorrelation behaviour persisting in a non stationary time series. A higher value of h(2) indicates a long range autocorrelation persisting in a non linear and non stationary time series where as the lower value of hurst exponent indicates that the persistence property decreases and the time series tends towards random brownian motion behaviour. As stated earlier in section-3, the multifractal spectrum i.e. variation of $f(\alpha)$ versus $\alpha$, resembles a wide inverted parabola, the feature $F_2$ indicates the value of value of the singularity exponent corresponding to maximum fluctuation of a time series. $\alpha_{peak}$ is the value of singularity exponent for which the singularity spectrum $f(\alpha)$ has a maxima. $\alpha_{peak}$ indicates the degree of correlation of a time series. A higher value of $\alpha_{peak}$ indicates that the data points are highly correlated. To elucidate further, if a past EEG signal emitted from human brain reveal spike, the probability of occurrence of the spike in the next EEG signal is greater than 0.5. The process is repetitive in nature and indicates a high degree of correlation between two pulses. On the other hand, if the subsequent EEG signals are not affected by spikes, and become independent of the previous state, then it indicates a lower degree of correlation and almost a regular pattern. $F_3$ and $F_4$ are the two extreme values of singularity exponent $\alpha$, which indicates the maximum and minimum fluctuations, respectively. Feature $F_5$ represents the mean of two extreme values of $\alpha$, which corresponds to average fluctuations. Feature $F_6$ is the width of the singularity spectrum. A time series showing high degree of fluctuations are characterized by a large value of spectral width. $F_7$ and $F_8$ denotes the horizontal distances between the peak and the minimum and maximum values of singularity exponents, respectively. This horizontal distance is the measure of the difference between average fluctuations with the minimum and maximum fluctuations of a time series. Features $F_9$ and $F_{10}$ are the ordinates i.e. singularity spectrum values $f(\alpha)$, corresponding to two extreme values of singularity exponents and $F_{11}$ represents the mean of the two. Feature $F_{12}$ is the vertical difference between $F_9$ and $F_{10}$. $F_9$ and $F_{10}$ indicate the unit number maximum and minimum probability subset in EEG signals and $F_{12}$ is the measure the proportion of large and small peaks in EEG signals [24]. For $F_{12} < 0$, the proportion of larger peaks in EEG signals are less compared to smaller peaks, hence amplitude distortion is lower. For $F_{12} > 0$, the proportion of large peaks are higher than small peaks, which indicates a higher amplitude distortion. Features $F_{13}$ and $F_{14}$ indicates the difference in height between the mean and the extreme values minimum and maximum) of the singularity spectrums, respectively.

## 5. Support vector machines

SVM is a supervised machine learning algorithm developed primarily to solve a binary classification problem. Detailed description of the SVM algorithm can be found out in [3,26]. An SVM performs classification by finding an optimum hyper plane having a maximum margin (i.e. the distance between the boundary and the nearest points) between the two classes using the principle of Structural Risk Minimization (SRM) [3]. In the present study since, all classification problems are binary in nature, therefore SVM classifier is used. In the present study since, all classification problems are binary in nature, therefore SVM classifier is used. In case of non linear SVMs, the training data are mapped into a high dimensional feature space using different kernel functions, which performs this mapping operation satisfying Mercer's theorem. There can be several kernel functions in an SVM like Linear, Polynomial, Radial Basis Function etc. In the initial part of the present work different kernel functions of SVM have been employed to test the performance of the classifier, and it has been observed that the performance of Radial Basis Function kernel has been found to be satisfactory for all cases. Mathematically, for a linearly separable training data (c, d) the RBF kernel function $\kappa(c,d)$ can be expressed as

$$\text{Radial basis function: } \kappa(c,d) = e^{(-\gamma\|c-d\|^2)} \quad (12)$$

where $\gamma$ is known as kernel parameter and $\gamma = \dfrac{1}{2\omega^2}$, with '$\omega$' as the width.

## 6. Results and Discussion

*6.1 Analysis of EEG signals using MFDFA:*

As highlighted in this paper, the EEG signals representing healthy, interictal and seizure activities are atfirst characterized by mutifractal parameters. Figure (a) shows the variation of generalized Hurst exponent against q for five different sets A-E. In the present work, 'q' is varied from -5 to +5 in steps of 0.1and the value of 's'(scale) is chosen between 16–1024, having a total number of 19 equal logarithmic intervals in between.

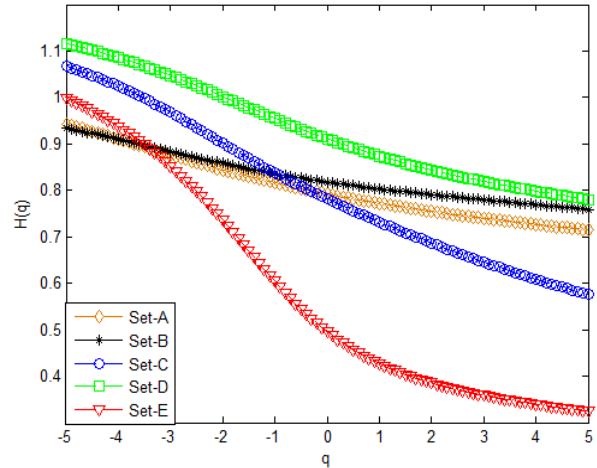

Figure1.Variation of h(q) vs. q for EEG signals of different sets

It can be observed from Figure 1, that the Hurst exponent curves shows a non linear relationship with 'q' for all sets indicating a multifractal nature of EEG signals. Moreover, since EEG signals of different sets represent different states of human brain, a wide variation in shape, size and position of Hurst exponent curves are observed in Figure 1. Further, since the smaller values of generalized Hurst exponent indicate large fluctuations for positive values of 'q', it is therefore evident from Figure 1, that epileptic seizure EEG signals corresponding to set-E, the generalized Hurst exponent shows minimum value for 'q' > 0 compared to inter-ictal and healthy EEG signals, which indicates high degree of fluctuations. Therefore, the Hurst exponent curves obtained for different EEG signals clearly bears the evidence of chaotic and non linear behavior representing complex dynamics of human brain. Figure 2 shows the variation of multifractal scaling exponent τ(q) versus q for different sets of EEG signals, which also shows a typical non linear behavior.

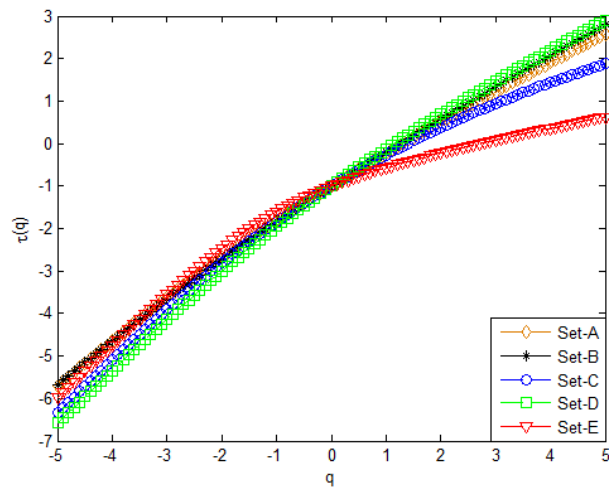

Figure 2.Variation of τ(q) vs. q for EEG signals of different sets

The shape of the curves are typically convex in nature, which clearly manifests a multifractal nature of EEG signals of different sets. The most interesting observation is the degree of non linearity, which is found to be highest for seizure signals corresponding to set-E, compared to healthy and inter-ictal EEG signals, respectively. which again indicates that during seizure activity, the EEG signals show a greater amount of fluctuations. Figure 3 shows the multifractal spectrum obtained for different EEG signals. It can be pointed out from Figure 3, that the MFS of different EEG signals reveal a wide inverted parabolic nature, with different values of singularity spectrum widths Δα. The greatest width of the MFS is obtained for seizure signals followed by interictal and healthy signals. Since, the width of MFS indicate a higher degree of multifractality, it is evident that during epileptic seizures, the EEG signals manifest a high degree of multifractality followed by interictal and healthy states. Moreover, it can be observed that the seizure EEG signal have extended 'right tail' characteristic which are absent for either healthy and seizure signals. MFS having extended 'right tail' characteristic indicate that the MFS are insensitive to local fluctuations with larger magnitudes. Besides, it can be observed that the MFS of different EEG signals are not perfectly symmetric, i.e. their singularity spectrums f(α) do not attain peak for a fixed value of singularity exponent α for all cases. Therefore different feature parameters can be extracted from their respective MFS to distinguish between different types of EEG signals. As mentioned earlier in Section-4 fourteen different features have been extracted initially, among which a feature ranking test is being done to select the most significant features for performing the classification task. The details of the selected features and feature ranking are discussed in the following section.

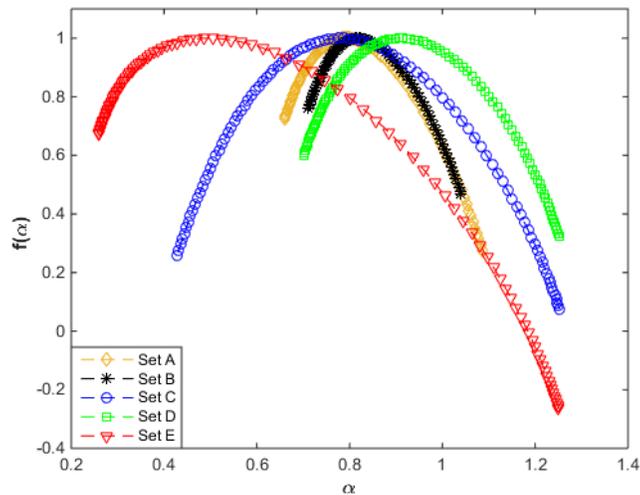

Figure 3.Variation of f(α) vs. α for EEG signals of different sets

*6.2 Feature Ranking using student's t-test:*

In the present study, a student's *t*-test is done to rank the features extracted from MFS of different EEG signals. The purpose of using a student's *t*-test is to reduce the size of the feature vector to eliminate feature redundancy and at the same time to improve the computational cost. In a student's *t*-test, the features are ranked on the basis of their *t*-values. A higher value of *t* indicate a better rank of a feature. For eight classification problems, eight paired student's *t*-test are conducted. For a two class problem like the present case, the outcome of the *t*-test yields a '*p*' value which is almost similar like a one way Analysis of Variance (ANOVA) test [28-29]. A lower '*p*' value indicates very high discrimination ability of the selected features . After conducting student's *t*-test, for each classification problem, sequential feature selection procedure (SFS) is adopted to determine the most optimal feature set for each classification problem. In SFS technique, a subset of features from the entire feature data set are selected sequentially according to the rank of the feature. The classification accuracy is tested each time with the selected feature, till no further improvement in accuracy is observed. The number of selected features for each CP is therefore the optimized feature set that can be used to train a classifier yielding maximum classification accuracy. The optimized features with their feature values and respective '*p*' values for eight CP are shown in Tables 2-9. It can be pointed out from Tables-2-9, that the selected optimal feature parameters for eight CP have a significant amount of class separation between them. Hence, the discriminative ability of the selected features are statistically significant upon statistical hypothesis testing. The performance parameters for each CP have been evaluated based on the selected optimal feature sets which are discussed in the following section.

Table-2: Results of paired students *t*-test for classification problem-I with feature values

| No of selected features | Selected Features | Feature values (Mean±Standard deviation) | | '*p*' values |
|---|---|---|---|---|
| | | A | E | |
| 3 | $F_1$ | 0.78 ± 0.08 | 0.41 ± 0.17 | 2.84e-37 |
| | $F_4$ | 0.68 ± 0.10 | 0.31 ± 0.14 | 9.94e-40 |
| | $F_{12}$ | 0.23 ± 0.14 | 0.62 ± 0.21 | 2.50e-26 |

Table-3: Results of paired students *t*-test for classification problem-II with feature values

| No of selected features | Selected Features | Feature values (Mean±Standard deviation) | | 'p' values |
|---|---|---|---|---|
| | | B | E | |
| 5 | $F_1$ | 0.62±0.10 | 0.41± 0.17 | 4.03e-18 |
| | $F_4$ | 0.54±0.12 | 0.31±0.14 | 2.32e-22 |
| | $F_6$ | 0.37±0.11 | 0.66±0.27 | 1.19e-16 |
| | $F_7$ | -0.26±0.07 | -0.48±0.18 | 1.36e-19 |

Table-4: Results of paired students *t*-test for classification problem-III with feature values

| No of selected features | Selected Features | Feature values (Mean±Standard deviation) | | 'p' values |
|---|---|---|---|---|
| | | C | E | |
| 4 | $F_1$ | 0.78± 0.08 | 0.41± 0.17 | 4.42e-36 |
| | $F_4$ | 0.87±0.08 | 0.49± 0.20 | 2.48e-32 |
| | $F_6$ | 0.21±0.14 | 0.62± 0.21 | 3.45e-24 |
| | $F_7$ | 0.53±0.13 | 0.27±0.15 | 1.95e-24 |

Table-5: Results of paired students *t*-test for classification problem-IV with feature values

| No of selected features | Selected Features | Feature values (Mean±Standard deviation) | | 'p' values |
|---|---|---|---|---|
| | | D | E | |
| 9 | $F_1$ | 0.73±0.11 | 0.41± 0.17 | 3.13e-29 |
| | $F_2$ | 0.84±0.16 | 0.49± 0.20 | 2.29e-23 |
| | $F_3$ | 1.24±0.22 | 0.97±0.35 | 6.16e-09 |
| | $F_4$ | 0.53±0.10 | 0.31±0.14 | 3.45e-24 |
| | $F_5$ | 0.89±0.13 | 0.64±0.23 | 3.64e-15 |
| | $F_8$ | 0.30±0.15 | 0.18±0.10 | 2.09e-09 |
| | $F_{10}$ | 0.48±0.16 | 0.74±0.16 | 2.25e-19 |
| | $F_{12}$ | 0.29±0.17 | 0.62± 0.21 | 1.95e-24 |

Table-6: Results of paired students *t*-test for classification problem-V with feature values

| No of selected features | Selected Features | Feature values (Mean±Standard deviation) | | 'p' values |
|---|---|---|---|---|
| | | AB | E | |
| 8 | $F_1$ | 0.68±0.13 | 0.41± 0.17 | 1.07e-24 |
| | $F_2$ | 0.72±0.13 | 0.49± 0.20 | 1.72e-17 |
| | $F_4$ | 0.59±0.13 | 0.31±0.14 | 1.23e-27 |
| | $F_6$ | 0.37±0.11 | 0.66±0.27 | 7.50e-17 |

| | F$_7$ | -0.24±0.08 | -0.48±0.18 | 5.50e-22 |
| | F$_9$ | 0.43±0.16 | 0.12±0.24 | 4.47e-19 |
| | F$_{12}$ | 0.29±0.19 | 0.62± 0.21 | 3.13e-19 |
| | F$_{13}$ | 0.57±0.16 | 0.89±0.24 | 4.33e-19 |

Table-7: Results of paired students *t*-test for classification problem-VI with feature values

| No of selected features | Selected Features | Feature values (Mean±Standard deviation) | | 'p' values |
|---|---|---|---|---|
| | | CD | E | |
| 7 | F$_1$ | 0.75±0.11 | 0.27± 0.15 | 1.68e-31 |
| | F$_2$ | 0.86±0.15 | 0.41± 0.17 | 3.94e-27 |
| | F$_4$ | 0.56±0.11 | 0.49± 0.20 | 2.90e-25 |
| | F$_5$ | 0.90 ±0.12 | 0.64±0.23 | 1.20e-17 |
| | F$_{10}$ | 0.47±0.15 | 0.74±0.16 | 1.09e-21 |
| | F$_{12}$ | 0.25±0.16 | 0.62± 0.21 | 5.30e-28 |
| | F$_{14}$ | 0.53±0.15 | 0.27± 0.15 | 7.58e-22 |

Table-8: Results of paired students *t*-test for classification problem-VII with feature values

| No of selected features | Selected Features | Feature values (Mean±Standard deviation) | | 'p' values |
|---|---|---|---|---|
| | | AB | CD | |
| 6 | F$_3$ | 0.96±0.12 | 1.24±0.20 | 1.30e-24 |
| | F$_6$ | 0.37±0.11 | 0.68±0.20 | 3.95e-25 |
| | F$_8$ | 0.13±0.06 | 0.30±0.13 | 5.27e-22 |
| | F$_{10}$ | 0.72±0.13 | 0.47±0.15 | 6.39e-24 |
| | F$_{11}$ | 0.58±0.10 | 0.36±0.12 | 2.54e-24 |
| | F$_{14}$ | 0.28±0.12 | 0.53±0.15 | 4.63e-24 |

Table -9: Results of paired students *t*-test for classification problem-VIII with feature values

| No of selected features | Selected Features | Feature values (Mean±Standard deviation) | | 'p' values |
|---|---|---|---|---|
| | | ABCD | E | |
| 7 | F$_1$ | 0.73±0.12 | 0.27± 0.15 | 3.67e-27 |
| | F$_2$ | 0.79±0.14 | 0.41± 0.17 | 5.81e-21 |
| | F$_4$ | 0.59±0.13 | 0.49± 0.20 | 4.19e-25 |
| | F$_7$ | -0.31±0.11 | -0.48±0.18 | 1.65e-11 |
|   | F9  | 0.35±0.17  | 0.12±0.24  | 1.42e-10 |
|   | F12 | 0.28±0.15  | 0.62± 0.21 | 3.89e-23 |
|   | F13 | 0.65±0.17  | 0.89±0.24  | 3.64e-11 |

*6.3 Performance Analysis of SVM classifier*:

The performance metric of the proposed seizure detection scheme is being evaluated using different statistical parameters like Accuracy, Senstivity, and Specificity. Mathematically, these parametes can be expressed as

$$Accuracy = (\frac{TP+TN}{TN+FN+TP+FP}) \times 100 \quad (13)$$

$$Sensitivity = (\frac{TP}{TP+FN}) \times 100 \quad (14)$$

$$Specificity = (\frac{TN}{TN+FP}) \times 100 \quad (15)$$

In the above equations, True Positive (TP), False Negative (FN), False Positive (FP) and True Negative (TN) are evaluated from the respective confusion matrix for all eight classification problems. True Positive and True Negative signifies the number of correctly classified cases and on the other False positive and False negative signifies the number of misclassified cases. In the present study, seizure signal is considered as negative class, whereas healthy and interictal are considered to be positive classes, respectively. Table-10 presents the performance parameters evaluated for eight CPs using SVM classifier. In case of SVM, the kernel functions are varied and it has been observed that the highest classification accuracy is obtained for Radial Basis Function (RBF) kernel compared to other kernel functions. Hence, performance parameters based on RBF kernel function have been reported in this paper. To assess the reliable performance of the classifiers, a tenfold cross validation technique has been adopted in this work. and at the same time to increase the robustness of the work. The value of kernel parameter $\gamma$ of the RBF kernel function in equation (13), should be optimized meticulously, since it can affect the classification accuracy significantly. The value of kernel parameter are generally selected using either a grid search algorithm or by implementing any other optimization algorithm like PSO, GA etc. In the present work, a grid search algorithm has been employed to find the optimal value of $\gamma$ yielding highest classification accuracy.

Table 10: Performance of the proposed method using SVM classifier

| CP | Acc | Sen | Spe |
|----|-----|-----|-----|

| CP | | | |
|---|---|---|---|
| I (A,E) | 100 | 100 | 100 |
| II (B,E) | 98.75 | 100 | 97.56 |
| III (C,E) | 100 | 100 | 100 |
| IV (D,E) | 100 | 100 | 100 |
| V (AB,E) | 100 | 100 | 100 |
| VI (CD,E) | 100 | 100 | 100 |
| VII (AB,CD) | 95.50 | 94.75 | 95.20 |
| VIII (ABCD,E) | 100 | 100 | 100 |

As it can be observed from Table-10, that the maximum classification accuracy of 100% is obtained for six cases out of eight CPs addressed in this paper. For CP-I, III, IV,V,VI and VIII, the proposed method yielded maximum accuracy of 100%. For CP-II and CP-VII, maximum accuracy of 98.75% and 95.50% have been achieved in this work. Besides, the maximum sensitivity and specificity of 100% has been obtained for seven and six CPs , respectively which is a significant improvement in comparison with the existing results. Therefore it can be said that the proposed method is highly sensitive and can also discriminate healthy, interical and seizure free EEG signals from seizure signals with utmost accuracy.

*6.3 Performance analysis using different classifiers:*

In order to ensure the robustness of the work, along with SVM, the performance of the proposed method is also being evaluated using different classifiers like k nearest Neighbour (kNN), Decision Tree (DT) and Probablistic Neural Network (PNN). Table-11 report the classification accuracies obtained for eight CPs using SVM, kNN, DT and PNN classifiers.

Table-11: Performance analysis using different classifiers

| CP | SVM | kNN | DT | PNN |
|---|---|---|---|---|
| I | 100 | 100 | 100 | 100 |
| II | 98.75 | 96.25 | 97.52 | 98.5 |
| III | 100 | 100 | 98.85 | 100 |
| IV | 100 | 98.45 | 100 | 98.72 |
| V | 100 | 100 | 99.15 | 100 |
| VI | 100 | 97.45 | 98.25 | 97.5 |
| VII | 95.50 | 92.75 | 93.50 | 94.25 |
| VIII | 100 | 98.25 | 97.75 | 100 |

It can be observed from Table-11, that 100% classification accuracy is obtained for six CPs using SVM classifier, followed by PNN which results in 100% classification accuracy for four CPs. The performance of kNN and DT classifiers are also reasonable satisfactory for all CPs, yielding 100% classification accuracy for three and two CPs respectively. However, the most important observation is that all classifiers have delivered consistent performance in classifying different EEG signals based on MFDFA based features for all eight CPs, which is an indicator of the stable and reliable performance of the proposed work.

*6.4 Comparative study with existing literatures:*

In this section, the performance of the proposed method employing MFDFA based feature extraction technique is compared with some state of the art techniques of seizure detection. Table-12 compare the classification accuracies obtained using the proposed method for CP-I-VIII, with the existing literatures studied on the same dataset, but using different methodology. It can be observed that the proposed method is capable of delivering almost identical and for some cases even better performance in comparison with some existing results for all classification problems addressed in this paper. For CP-II, the results presented in [18] is higher than the present work, but the method proposed in [18] used more number of feature sets as compared to the present work. For the rest of the CPs, the proposed method is found to outperform most of the recently published results. Hence, the proposed method based on MFDFA based features has a reasonably high degree of accuracy in detecting healthy, interictal and seizure free signals from epileptic seizure EEG signals and can be applied for clinical diagnosis of patients suffering from epilepsy.

Table-12: Comparative study with state of the art methods

| CP | Method | Accuracy (%) |
|---|---|---|
| I (A,E) | Chandaka et al., [3] | 95.96 |
| | Kaya et al., [12] | 99.50 |
| | Sammie et al.,[31] | 99.80 |
| | Supriya et al.,[17] | 100 |
| | Swami et al., [30] | 100 |
| | Sharma et al.,[18] | 100 |
| | **Proposed Work** | **100** |
| II | Nicoletta et al., [32] | 82.9 |
| (B,E) | Supriya et al., [17] | 97.25 |
| | Sharma et al.; [18] | 100 |
| | **Proposed Method** | **98.75** |
| III<br>(C,E) | Sammie et al., [31] | 98.50 |
| | Supriya et al.,[17] | 98.50 |
| | Swami et al., [30] | 98.72 |
| | Sharma et al; [18] | 99.00 |
| | **Proposed Method** | **100** |
| IV<br>(D,E) | Supriya et al.,[17] | 93.25 |
| | Kaya et al., [12] | 95.5 |
| | Swami et al., [30] | 93.33 |
| | Sharma et al., [18] | 98.50 |
| | **Proposed Method** | **100** |
| V<br>(AB, E) | Swami et al., [30] | 99.18 |
| | Sharma et al., [18] | 100 |
| | **Proposed Method** | **100** |
| VI<br>(CD,E) | Swami et al., [30] | 95.15 |
| | Kaya et al. [12] | 97.00 |
| | Sharma et al., [18] | 98.67 |
| | **Proposed Method** | **100** |
| VII<br>(AB,CD) | Sharma et al., [18] | 92.50 |
| | **Proposed Method** | 95.50 |
| VIII<br>(ABCD,E) | Sammie et al.,[31] | 98.1 |
| | Swami et al., [30] | 95.24 |
| | Sharma et al., [18] | 100 |
| | **Proposed Method** | **100** |

**Conclusion**

In this paper, a novel feature extraction technique based on MFDFA is presented for automated detection and classification of EEG signals. The EEG signals representing different states of human brain (healthy, inter-ictal and seizure) are at first analysed using MFDFA. It has been observed that the MFS of EEG signals during healthy, inter-ictal and seizure states of brain behave differently indicating a wide variation in their nature of multifractality. From the MFS of five EEG signals, fourteen new features have been extracted to discriminate between different types of EEG signals. After feature ranking employing student's *t*-test, features with high discriminative ability have been selected using SFS technique to serve as input feature sets for effective classification of EEG signals. SVM classifier have been implemented in the present work and its performance is also compared with several benchmark classifiers. Besides, eight different classification problems have been reported in this study and it has been observed that the proposed method is capable of yielding 100% classification accuracy in six cases indicating the reliability of the proposed work. The performance of the proposed MFDFA aided SVM classifier is also found to outperform the existing methods in terms of overall classification accuracy for many classification problems. Hence, it can be inferred that the proposed method can be potentially implemented in practice for diagnosis of epilepsy. The present method has an added advantage of lower computational burden since MFDFA is computationally inexpensive compared to several other non linear analysis techniques. However, the present analysis is based on a single channel available EEG signal recording consisting of only 4097 samples. It would be interesting to observe the efficacy of the proposed method when it is being applied to large EEG recordings and especially for multiple channels, which will be done as a part of the future research work. The present work employing MFDFA based feature extraction technique will also be applied in future in the field of automated diagnosis of not only epilepsy but also several other neurological and neuromuscular disorders etc.

.